\documentclass[11pt]{article}

\usepackage[margin=1in]{geometry}
\usepackage[T1]{fontenc}
\usepackage{amsmath}
\usepackage{amssymb}
\usepackage{booktabs}
\usepackage{array}
\usepackage{longtable}
\usepackage{hyperref}
\usepackage{listings}
\usepackage{xcolor}
\usepackage{parskip}
\usepackage{enumitem}
\usepackage{textcomp}
\usepackage{graphicx}
\usepackage{float}

\hypersetup{
  colorlinks=true,
  linkcolor=blue,
  urlcolor=blue,
  citecolor=blue
}

\newcommand{\code}[1]{\texttt{#1}}

\title{\textbf{reward-lens: A Mechanistic Interpretability Library for Reward Models}}
\author{Mohammed Suhail B. Nadaf\thanks{Correspondence: \texttt{suhailnadaf509@gmail.com}.}}
\date{}

\begin{document}

\maketitle

\hrule
\vspace{1em}

\begin{abstract}
\noindent
Every RLHF-trained language model is shaped by a reward model, yet the
mechanistic-interpretability toolkit has been built almost entirely for
generative LLMs. Its primitives --- the logit lens, direct logit
attribution, attribution patching, ACDC-style circuit discovery, sparse
autoencoders --- all terminate in a projection onto the vocabulary
unembedding $W_U$, which a reward model replaces with a scalar
regression head. We present \code{reward-lens}, an open-source library
that ports this toolkit to reward models. The library is organised
around one observation: the reward head's weight vector $w_r$ is the
natural axis along which every interpretability question for a reward
model can be posed and decomposed. It packages a Reward Lens (the
reward-model analogue of the logit lens), component attribution,
contrastive activation patching, a reward-hacking probe suite, and TopK
sparse-autoencoder feature attribution against $w_r$, alongside five
extension modules grounded in recent 2025--2026 alignment-theory
results: a distortion index, divergence-aware patching, a misalignment
cascade detector, a reward-term conflict analyser, and concept-vector
analysis. We validate the framework on two production reward models
(Skywork-Reward-Llama-3.1-8B-v0.2 and ArmoRM-Llama3-8B-v0.1) across
$\sim$695 RewardBench preference pairs per model (1390 pair-model
evaluations across the two reward models). The central empirical finding
is negative: linear attribution does not predict causal importance
(mean Spearman $\rho = -0.256$ on Skywork, $-0.027$ on ArmoRM). We
treat this as a property the framework is designed to expose, not a
bug, and argue that observational decomposition and causal intervention
should be first-class and easily compared.
\end{abstract}

\vspace{1em}
\hrule
\vspace{1em}

\section{Introduction}

Reinforcement learning from human feedback has become the standard
training recipe for frontier language models, and every model trained
this way has its behaviour shaped by a learned scalar function --- the
reward model --- that stands in for human preferences during the
policy-optimisation stage. The reward model is therefore the most
safety-critical component of the alignment pipeline: if it encodes the
wrong preferences, an optimised policy inherits those preferences
exactly, often amplified.

The community's response to this fact has been asymmetric. Behavioural
evaluation of reward models is a mature cottage industry: RewardBench
(Lambert et al., 2024), HelpSteer (Wang et al., 2024), and a long
literature on specific failure modes (length bias, sycophancy,
formatting bias, position bias). Mechanistic evaluation of reward
models is almost non-existent. The tools that power generative-LLM
interpretability --- TransformerLens, the logit lens, direct logit
attribution, attribution patching, ACDC, and the SAE ecosystem --- all
project intermediate model states onto the vocabulary unembedding $W_U$
and score them against a target token distribution. Reward models
replace that unembedding with a one-dimensional linear head, which
silently breaks each primitive. The gap is structural, not incidental.

This paper introduces \code{reward-lens}, a library that fills the
gap, and documents its first systematic application to two open-weight
production reward models. The library is built around one organising
observation: the reward head is $r = w_r^\top h_\mathrm{final} + b_r$,
so $w_r$ is the natural axis along which every interpretability
question for a reward model should be posed. Projecting an intermediate
residual stream onto $w_r$ yields the \emph{Reward Lens}, the analogue
of the logit lens. Decomposing the residual stream into per-component
contributions and projecting each onto $w_r$ yields the analogue of
direct logit attribution. Swapping component activations between
preferred and dispreferred completions and measuring the change in
$w_r^\top h_\mathrm{final}$ yields activation patching with a
contrastive structure built into the setup --- something the
generative-model case does not have for free.

\paragraph{Contributions.}
(i) \emph{A library.} \code{reward-lens} is the first installable,
architecture-agnostic toolkit for mechanistic interpretability of
reward models, with a small adapter protocol validated on Llama,
Mistral, Gemma-2, and ArmoRM-style multi-objective heads and a generic
auto-detecting adapter for any HuggingFace
\code{AutoModelForSequenceClassification} reward model. The library
is the principal contribution of this paper; the experiments are
validation (source: \url{https://github.com/suhailnadaf509/reward-lens}).
(ii) \emph{A unifying framing.} Every useful interpretability primitive
for a reward model is a projection onto, or decomposition along, $w_r$.
This unifies the lens, attribution, patching, SAE-feature analysis, and
concept-vector analysis into a single family with shared algebra and
shared plotting utilities.
(iii) \emph{Five theory-grounded extensions} (distortion index,
divergence-aware patching, misalignment cascade, reward-term conflict,
concept-vector analysis) that operationalise recent alignment-theory
results in runnable code; the contribution of \code{reward-lens} is the
operationalisation, not the theory.
(iv) \emph{Empirical validation} on
Skywork-Reward-Llama-3.1-8B-v0.2 and ArmoRM-Llama3-8B-v0.1 over
$\sim$695 RewardBench pairs spanning crystallisation depth,
attribution structure, hacking profiles, reward-term geometry, circuit
overlap, concept dose-response, and cross-model formation correlations.
(v) \emph{A negative result.} Linear attribution does not predict
causal patching effects in either model; the disagreement is sharper
here than in the generative-model case, plausibly because a scalar
output leaves more room for redundant pathways to compensate. The
library keeps observational and causal views first-class and trivially
comparable on the same set of components.

\vspace{1em}
\hrule
\vspace{1em}

\section{Background}

Five years of interpretability work on decoder-only transformers has
converged on a small family of primitives: the logit lens
(nostalgebraist, 2020) and tuned lens (Belrose et al., 2023) project
intermediate residual streams through the unembedding $W_U$; direct
logit attribution decomposes the residual stream into per-sublayer
contributions and projects each onto $W_U$; attribution patching (Syed
\& Nanda, 2024) and ACDC (Conmy et al., 2023) extend this with
linearised and search-based causal discovery; the SAE ecosystem (Bloom,
2024; Gao et al., 2024; Rajamanoharan et al., 2024) decomposes residual
streams into dictionary features. Every one of these primitives ends in
a projection onto $W_U$ and a comparison against a target token
distribution. For a reward model, the head is not $W_U$; it is a single
linear functional $w_r$, and the off-the-shelf primitives need to be
re-derived.

A reward model is trained to output a scalar quality score $r(x, y)$ on
a Bradley--Terry preference loss (Christiano et al., 2017; Ouyang et
al., 2022). Production reward models include single-scalar-head models
(Skywork-Reward-Llama-3.1-8B-v0.2, Liu et al., 2024; FsfairX),
multi-objective models with prompt-conditional MoE gating (ArmoRM, Wang
et al., 2024), and multi-dimensional-head models
(Nemotron-4-340B-Reward). Behavioural characterisation of their failure
modes is rich; mechanistic characterisation is rare. This is not
because reward-model interpretability is harder than the generative
case. In several structural respects it is easier: the target of
attribution is a single number rather than a $50{,}000$-way
distribution; the contrastive structure of preference data is native;
and the projection axis $w_r$ is fixed by construction rather than
chosen. \code{reward-lens} is our attempt to exploit those advantages.

\vspace{1em}
\hrule
\vspace{1em}

\section{The $w_r$-centred view}

Throughout we treat the final layer norm as absorbed into $w_r$ (we
extract $w_r$ from the score-head weight and keep the norm in the
forward pass). Let $h^{(\ell)} \in \mathbb{R}^d$ denote the residual
stream at the final token position after layer $\ell$, with $\ell = -1$
the embedding output and $\ell = L-1$ the output of the final
transformer block. Let $\mathrm{attn}^{(\ell)}$ and
$\mathrm{mlp}^{(\ell)}$ denote the per-sublayer residual contributions
added into the stream at layer $\ell$. The reward head is
\begin{equation}
r(x, y) \;=\; w_r^\top h_\mathrm{final}(x, y) \;+\; b_r,
\end{equation}
and the residual stream decomposes additively across sublayers:
\begin{equation}
h_\mathrm{final} \;=\; h_\mathrm{embed} \;+\; \sum_{\ell=0}^{L-1}\Bigl(
\mathrm{attn}^{(\ell)} + \mathrm{mlp}^{(\ell)} \Bigr).
\end{equation}
Pulling $w_r$ through the sum gives an exact per-component decomposition
of the scalar reward:
\begin{equation}
r \;=\; w_r^\top h_\mathrm{embed} \;+\; \sum_{\ell=0}^{L-1}\Bigl(
w_r^\top \mathrm{attn}^{(\ell)} + w_r^\top \mathrm{mlp}^{(\ell)} \Bigr) + b_r.
\end{equation}

Equation~(3) is the decomposition every primitive in \code{reward-lens}
is built on. \emph{Reward Lens:} project the intermediate residual
stream at layer $\ell$ onto $w_r$,
\begin{equation}
\mathrm{RewardLens}^{(\ell)}(x, y) = w_r^\top h^{(\ell)}(x, y) + b_r,
\end{equation}
and for a preference pair report the differential
$\Delta^{(\ell)} = \mathrm{RewardLens}^{(\ell)}(\text{pref}) - \mathrm{RewardLens}^{(\ell)}(\text{dispref})$.
\emph{Component attribution:} compute each term of (3) and report it as
a signed scalar (and its preference-pair differential).
\emph{SAE-feature decomposition:} given a TopK sparse autoencoder
reconstructing $h = \sum_i f_i d_i + b_\mathrm{dec}$, the reward
decomposes as
\begin{equation}
r \;\approx\; \sum_i f_i (w_r^\top d_i) + w_r^\top b_\mathrm{dec} + b_r,
\end{equation}
so $w_r^\top d_i$ is each feature's reward alignment --- a property of
the SAE that needs no forward pass to evaluate.
\emph{Activation patching:} swap $\mathrm{attn}^{(\ell)}$ or
$\mathrm{mlp}^{(\ell)}$ between source and target completions and
measure the induced change in $w_r^\top h_\mathrm{final}$.

\paragraph{The caveat the framing makes visible.} Equations (3) and (5)
are linear; the forward pass that produces $h_\mathrm{final}$ is not.
Layer norms, attention softmaxes, and residual-stream redundancy mean
that a component's \emph{contribution} to $h_\mathrm{final}$ (what (3)
measures) and the \emph{counterfactual effect} of ablating that
component (what patching measures) are different quantities. They can
and do disagree. Section~\ref{sec:experiments} quantifies the
disagreement on two production models. We do not treat this as a
failure of the framing; we treat it as a property of reward-model
computation that is made visible by the framing.

\vspace{1em}
\hrule
\vspace{1em}

\section{The library}
\label{sec:library}

\code{reward-lens} (\code{pip install reward-lens}, version 1.0.0)
wraps any HuggingFace \code{AutoModelForSequenceClassification} reward
model behind one Python object and exposes a small set of primitives
organised around the $w_r$-centred view of \S3. This section gives the
gist; depth lives in the appendix, where each primitive is paired with
its generative-LLM antecedent and its full algorithmic specification.

\subsection{Design principles}
\label{sec:design}

Four principles shape the library.

\emph{$w_r$ as the organising axis.} Every primitive is either a
projection onto $w_r$ (lens, component attribution, SAE feature
alignment, concept alignment) or a causal probe whose readout is the
change in $w_r^\top h_\mathrm{final}$ (activation patching,
divergence-aware patching, concept dose-response). The same vector
appears in every signature; the same plotting routines aggregate
results from every primitive.

\emph{HuggingFace-native, not TransformerLens-bound.} TransformerLens
hardcodes a vocabulary unembedding $W_U$ into the type of every
high-level primitive --- the precise assumption a reward model breaks.
Building on the HuggingFace \code{AutoModel} abstraction means
\code{reward-lens} can wrap any reward model someone has uploaded
without weight surgery, including models with non-standard heads
(multi-objective, MoE-gated). The cost is that we re-implement the
hook machinery; the benefit is coverage.

\emph{An adapter protocol for architecture-agnosticism.} Llama,
Mistral, Gemma-2, and ArmoRM all expose layers, attention sublayers,
and MLP sublayers under different attribute names; reward heads are
sometimes \code{score}, sometimes \code{classifier}, sometimes a
multi-tensor object behind a gating module. A ten-method adapter
protocol isolates these family-specific details from the rest of the
library, so the lens, attribution, patching, hacking, SAE, and
extension modules are written once against the protocol.

\emph{Observational and causal views, kept distinct.} A recurring
failure mode of interpretability work is to read off a contribution and
report it as if it were a cause. The library makes the difference
syntactic: lens and component attribution return observational
quantities (\code{ComponentResult}); the patcher returns a causal one
(\code{PatchingResult}); both have the same component schema, so a
single comparison plot is one line. \S\ref{sec:faithfulness} reports
what happens when one does this.

\subsection{The model wrapper and adapter protocol}
\label{sec:wrapper}

The wrapper, \code{RewardModel}, takes a HuggingFace model name or
path and returns an object exposing $w_r$ (\code{reward\_direction}),
$b_r$ (\code{reward\_bias}), the architectural sizes
($d_\mathrm{model}$, $n_\mathrm{layers}$, $n_\mathrm{heads}$,
$d_\mathrm{head}$), and a \code{forward\_with\_cache} method that runs
a forward pass while collecting per-layer residual streams and
per-sublayer attention and MLP outputs at the final token via PyTorch
forward hooks. An optional \code{cache\_full\_sequences} flag retains
the full sequence-length tensors needed for activation patching. Hooks
are registered as a context manager, so re-entrancy and cleanup are
explicit.

The adapter protocol abstracts the family-specific details into ten
methods covering reward-head extraction, layer iteration, sublayer
access, and output parsing for the various output formats HuggingFace
modules return (tuples, dataclasses, bare tensors). Concrete adapters
ship for Llama, Mistral, Gemma-2, and ArmoRM; a generic adapter
auto-detects layer and head names by walking the module tree and is
sufficient for most additional models without extra code. ArmoRM is
the stress test: its 19-objective regression head with
prompt-conditional MoE gating means there is no single ``reward
direction'' in the simple sense. The adapter exposes a default
\code{reward\_direction} (the mean of the 19 objective directions, the
right approximation for aggregate analyses) and a
\code{per\_objective\_directions} method returning the
$19 \times d_\mathrm{model}$ matrix for dimension-resolved analysis.
The full method-by-method contract and a per-family behaviour table
appear in Appendix~\ref{app:adapter}.

\subsection{Core primitives}
\label{sec:core}

Each primitive is the reward-model analogue of a generative-LLM
primitive. The mapping is the spine of the library and of the
appendix. The reader who knows the logit lens already knows what the
Reward Lens is the moment $W_U$ is replaced by $w_r$.

\paragraph{Reward Lens.} Project the intermediate residual stream at
each layer onto $w_r$ and report the per-layer trajectory of the
preference differential. This is the logit lens with $W_U$ replaced
by $w_r$ and the cross-entropy-against-a-target replaced by the scalar
preference differential. The result includes the per-layer
\emph{crystallisation depth}: the first fractional layer index at
which the differential reaches 50\,\% of its final value. Trajectory,
crystallisation, and marginal per-layer contributions are all returned
in a single \code{RewardLensResult}.

\paragraph{Component attribution.} Materialise equation~(3) by computing
$w_r^\top \mathrm{attn}^{(\ell)}$ and $w_r^\top \mathrm{mlp}^{(\ell)}$
for every $\ell$. This is direct logit attribution rephrased against
$w_r$. The result is a flat vector of $2L+1$ signed contributions
(plus the bias) with $\mathrm{top\_k}$, $\mathrm{by\_type}$, and
heat-map utilities. Because the reward target is a single scalar, the
output is one column per pair, not one column per vocabulary token ---
the bookkeeping is correspondingly cleaner.

\paragraph{Activation patching.} Swap a chosen sublayer's full-sequence
activation between two completions and re-run the forward pass,
measuring the change in $w_r^\top h_\mathrm{final}$. Three modes are
supported: \emph{noising} (target = preferred, source = dispreferred,
measures necessity), \emph{denoising} (the symmetric mode, measures
sufficiency), and \emph{zero ablation} (source replaced with zeros).
Generative-model patching has the same three modes; what is new for
reward models is that the contrastive structure is native --- a
preference pair is exactly the (clean, corrupted) input the patcher
expects. Returns a \code{PatchingResult} with the same component schema
as \code{ComponentResult}, so observational and causal views can be
plotted on the same axis.

\paragraph{Hacking detector.} Run paired A/B probes for length,
confidence, formatting, sycophancy, and repetition biases and report
Cohen's $d$ per dimension. This is the analogue of behavioural
red-teaming, but executed inside the library against a single reward
model rather than against an end-to-end RLHF policy. A user-supplied
custom probe set is the recommended workflow; the small shipped
defaults are diagnostic, not statistical.

\paragraph{TopK sparse-autoencoder feature attribution.} A TopK SAE
trained on a chosen residual layer reconstructs
$h \approx \sum_i f_i d_i + b_\mathrm{dec}$, after which equation (5)
gives a per-feature reward alignment $w_r^\top d_i$ that requires no
further forward pass to evaluate. This is the SAE-circuit-discovery
pipeline of Bloom (2024) and Rajamanoharan et al.\ (2024) with the
target replaced by $w_r$ instead of a token. The library ships the SAE
module, an activation collector that streams residual-stream
activations from a list of (prompt, response) inputs to disk, a trainer
with cosine-annealed Adam, and a feature analyser that returns
top/bottom features by reward alignment, activation frequency, and
top-activating examples.

\paragraph{Cross-model comparator.} Run the lens (and optionally
attribution) on a set of models for the same preference pair, normalise
the differential trajectories to common fractional depth, and report
crystallisation fractions plus pairwise Pearson correlations. The
\S5.4 cross-model formation overlap is one call.

\subsection{Theory-grounded extensions}
\label{sec:ext}

Five extension modules ground recent alignment-theory results in
runnable code. The original framework is due to the cited authors in
each case; the contribution of \code{reward-lens} is the
operationalisation. We sketch each here and give algorithmic detail in
Appendix~\ref{app:extensions}.

\paragraph{Distortion index.} Operationalises the
\emph{Reward Hacking as Equilibrium under Finite Evaluation} framework
(Wang \& Huang, 2026): given a set of quality dimensions and a set of evaluation
probes that target subsets of those dimensions, score each dimension's
\emph{effective coverage} from the probes' discriminative power and
return the under-coverage as a per-dimension distortion index. High
distortion means the dimension is not being measured; under
optimisation pressure it will be hacked.

\paragraph{Divergence-aware patching.} Operationalises
\emph{Addressing Divergent Representations from Causal Interventions
on Neural Networks} (2025, arXiv:2511.04638): wraps the activation patcher with a Mahalanobis-distance check
on the patched activation against an in-distribution covariance fitted
from clean preference data. Patches whose result lies far off
manifold are flagged, and divergences are classified as
\emph{harmless} (off-manifold but small effect, plausibly null-space)
or \emph{pernicious} (off-manifold with large effect, plausibly a
spurious activation of an unrelated circuit). The patcher returns the
same effect numbers as the unaware patcher plus a reliability score.

\paragraph{Misalignment-cascade detector.} Operationalises
\emph{Natural Emergent Misalignment from Reward Hacking} (MacDiarmid
et al., 2025): runs a battery of probes across six misalignment dimensions
(alignment-faking, malicious cooperation, sabotage, self-preservation,
deception, sycophancy), computes the across-dimension correlation
matrix of preference deltas, and reports systemic-risk clusters. High
correlation between dimensions means a single underlying mechanism is
driving multiple visible failure modes.

\paragraph{Reward-term conflict analyser.} Operationalises
\emph{Aligned, Orthogonal or In-Conflict} (Kaufmann et al., 2026):
given two or more reward terms (extracted from contrastive activations or supplied
directly), computes pairwise cosine similarities of the term
directions and classifies pairs as aligned ($\cos > 0.5$), orthogonal
($|\cos| < 0.2$), or in-conflict ($\cos < -0.3$). For multi-objective
heads such as ArmoRM, the analyser can be pointed directly at the
per-objective direction matrix.

\paragraph{Concept-vector analysis.} Following the methodology of
Chen et al.\ (2025) on concept directions in language models,
extracts linear concept directions from contrastive activations
(confidence, formality, agreement, verbosity, hedging, helpfulness as
shipped defaults; user-supplied concepts are the recommended
workflow), reports each concept's cosine alignment with $w_r$, and
produces a causal dose-response curve by adding $\alpha v$ into the
residual stream and measuring the change in reward. Concepts that are
both well-aligned with $w_r$ \emph{and} known to be hackable (e.g., a
length or sycophancy concept) carry a hacking-risk flag.

\subsection{A worked example}
\label{sec:worked}

The end-to-end use of the library on a single preference pair is short
enough to fit in a code block. The same calls aggregate trivially over
a dataset.

\begin{lstlisting}[language=Python]
from reward_lens import (
    RewardModel, RewardLens, ComponentAttribution,
    ActivationPatcher, HackingDetector,
)
from reward_lens import quick_concept_analysis

rm = RewardModel.from_pretrained("Skywork/Skywork-Reward-Llama-3.1-8B-v0.2")

# observational
trace      = RewardLens(rm).trace(prompt, preferred, dispreferred)
attribute  = ComponentAttribution(rm).attribute(prompt, preferred, dispreferred)

# causal
patch      = ActivationPatcher(rm).patch_all_components(
                 prompt, preferred, dispreferred, mode="noising")

# vulnerability profile
hacking    = HackingDetector(rm).scan(prompt=prompt, response=preferred)
concepts   = quick_concept_analysis(rm)

# observational vs causal, on the same components
trace.plot(); attribute.plot_top_k(10); patch.plot_top_k(10)
\end{lstlisting}

The principal output dataclasses (\code{RewardLensResult},
\code{ComponentResult}, \code{PatchingResult}, \code{HackingReport},
\code{ConceptAlignmentReport}, plus the extension reports) share a
small set of conventions documented in Appendix~\ref{app:api}: the
component schema is the same across attribution and patching; the
plotting utilities accept any of these dataclasses; aggregation across
a dataset is plain Python.

\vspace{1em}
\hrule
\vspace{1em}

\section{Experiments}
\label{sec:experiments}

\subsection{Setup}

We evaluate Skywork-Reward-Llama-3.1-8B-v0.2 (32 layers,
$d_\mathrm{model} = 4096$, single-scalar head) and ArmoRM-Llama3-8B-v0.1
(32 layers, multi-objective head with 19 objectives and
prompt-conditional MoE gating), both in bfloat16 on a single NVIDIA
A100-80GB. The preference data are RewardBench pairs ($n = 200$ each
for helpfulness, safety, correctness; $n = 95$ for verbosity;
$\sim$695 pairs per model, 1390 pair-model evaluations in total). Sanity check: the curated diagnostic preference set
is solved at 93.5\% accuracy by Skywork and 94.2\% by ArmoRM, so
neither model is confused on the controlled probes used by the hacking
detector and concept extractor. Reproducibility details appear in
Appendix~\ref{app:repro}.

\subsection{When does preference form?}

We run the Reward Lens on every pair, per dimension, for both models,
and record the fractional layer depth at which the differential first
reaches 50\% of its final value. Table~\ref{tab:cryst} reports the
per-pair mean and standard deviation.

\begin{table}[h]
\centering
\begin{tabular}{lcccc}
\toprule
& \multicolumn{2}{c}{Skywork} & \multicolumn{2}{c}{ArmoRM} \\
\cmidrule(lr){2-3}\cmidrule(lr){4-5}
Dimension    & Cryst.\ depth & $|\Delta_{\text{final}}|$ & Cryst.\ depth & $|\Delta_{\text{final}}|$ \\
\midrule
helpfulness  & $0.90 \pm 0.13$ & 6.4 & $0.73 \pm 0.26$ & 0.08 \\
safety       & $0.90 \pm 0.11$ & 9.4 & $0.78 \pm 0.22$ & 0.10 \\
correctness  & $0.91 \pm 0.11$ & 4.5 & $0.64 \pm 0.34$ & 0.02 \\
verbosity    & $0.93 \pm 0.06$ & 5.2 & $0.68 \pm 0.24$ & 0.05 \\
\bottomrule
\end{tabular}
\caption{Reward Lens crystallisation depth (fractional layer index,
mean $\pm$ std over $n = 200$ pairs per dimension; $n = 95$ for
verbosity) and mean final-differential magnitude in each model's
native reward units. Skywork crystallises uniformly late ($\sim$0.90);
ArmoRM crystallises earlier and with much higher across-pair variance.}
\label{tab:cryst}
\end{table}

Skywork crystallises preference uniformly late ($\sim$0.90 across all
four dimensions, small across-pair variance) --- consistent with a
single unified preference mechanism assembled near the output. ArmoRM
is messier: lower mean depths, higher across-pair variance (std
0.22--0.34), and a different ordering (correctness crystallises
earliest in the mean). We read this as a mechanistic signature of
ArmoRM's multi-objective architecture, exactly what a
prompt-conditional MoE-gated head would induce. The absolute reward
scales differ by $\sim$two orders of magnitude between models (Skywork
4--10 units; ArmoRM 0.02--0.10); all subsequent comparisons report
effects in each model's native units.

\begin{figure}[H]
  \centering
  \includegraphics[width=0.85\linewidth]{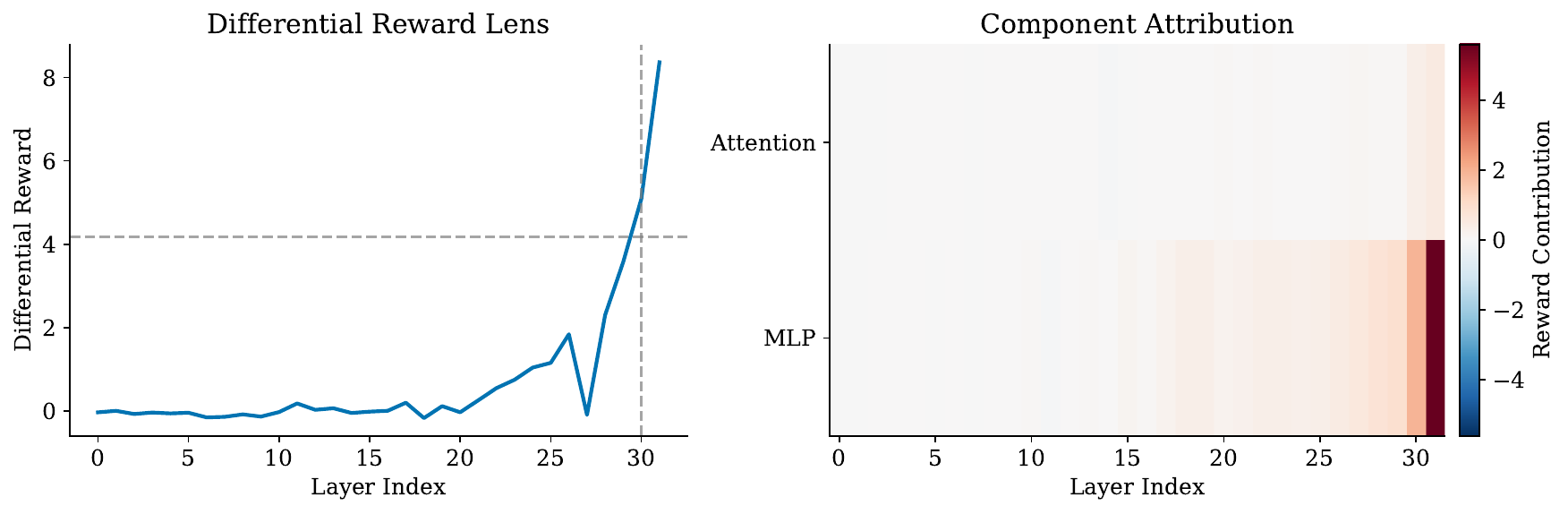}
  \caption{Reward Lens dashboard for a representative Skywork
  helpfulness pair: layer-wise preference formation and per-component
  attribution.}
  \label{fig:lens-dashboard}
\end{figure}

\subsection{Where does the model put its reward signal?}

Component attribution materialises equation (3). For each pair we rank
components by $|$contribution$|$; we summarise attribution structure by
the \emph{frequency} with which a component appears in the top-10
across the 200 pairs of a dimension (Table~\ref{tab:attrib}).

\begin{table}[h]
\centering
\begin{tabular}{llccc}
\toprule
Model   & Dimension     & Top 1 (freq.)            & Top 2 (freq.)            & Top 3 (freq.)            \\
\midrule
Skywork & helpfulness   & \code{mlp\_L31} (199/200)& \code{mlp\_L30} (196/200)& \code{mlp\_L29} (186/200)\\
Skywork & safety        & \code{mlp\_L31} (198/200)& \code{mlp\_L30} (197/200)& \code{mlp\_L29} (193/200)\\
Skywork & correctness   & \code{mlp\_L31} (199/200)& \code{mlp\_L30} (195/200)& \code{mlp\_L28} (193/200)\\
ArmoRM  & helpfulness   & \code{mlp\_L31} (196/200)& \code{mlp\_L30} (139/200)& \code{mlp\_L29} (112/200)\\
ArmoRM  & safety        & \code{mlp\_L31} (193/200)& \code{mlp\_L12} (139/200)& \code{attn\_L30} (101/200)\\
ArmoRM  & correctness   & \code{mlp\_L31} (192/200)& \code{mlp\_L30} (152/200)& \code{attn\_L30} (110/200)\\
\bottomrule
\end{tabular}
\caption{Top three components by attribution frequency across pairs.
Frequencies are out of 200 pairs. The last three MLPs
(\code{mlp\_L29-31}) dominate observational attribution
near-universally on Skywork; ArmoRM shows the same dominance for its
very last MLP but more dispersion below the top.}
\label{tab:attrib}
\end{table}

The dominance of \code{mlp\_L31} is the cleanest near-universality in
the data: the final MLP sublayer is in the top-10 for at least 192 of
200 pairs in every model/dimension cell. By observational attribution,
the reward signal in both models lives in the last three MLPs.

\subsection{Causal vs.\ observational: the faithfulness gap}
\label{sec:faithfulness}

We next ask whether component attribution predicts causal importance.
For the first preference pair in each dimension we run the noising
patcher on all 64 components (32 layers $\times$ \{attention, MLP\})
and record the Spearman rank correlation between $|{\rm attribution}|$
and $|{\rm patch\,effect}|$. Table~\ref{tab:faith} reports the result.

\begin{table}[h]
\centering
\begin{tabular}{llrr}
\toprule
Model   & Dimension     & Spearman $\rho$ & p-value  \\
\midrule
Skywork & helpfulness   &   $-0.445$   & 0.0002   \\
Skywork & correctness   &   $-0.255$   & 0.042    \\
Skywork & safety        &   $-0.067$   & 0.598    \\
Skywork & \textbf{mean} &   $-0.256$   &  ---     \\
ArmoRM  & helpfulness   &   $-0.282$   & 0.024    \\
ArmoRM  & correctness   &   $+0.125$   & 0.325    \\
ArmoRM  & safety        &   $+0.075$   & 0.556    \\
ArmoRM  & \textbf{mean} &   $-0.027$   &  ---     \\
\bottomrule
\end{tabular}
\caption{Spearman rank correlation between $|$attribution$|$ and
$|$patching effect$|$ across the 64 components of each preference pair.
$n = 64$ components; p-values are two-sided. No per-dimension
correlation exceeds $|0.3|$ on the positive side; three of six are
negative; the two significant correlations are both negative.}
\label{tab:faith}
\end{table}

\begin{figure}[H]
  \centering
  \includegraphics[width=0.6\linewidth]{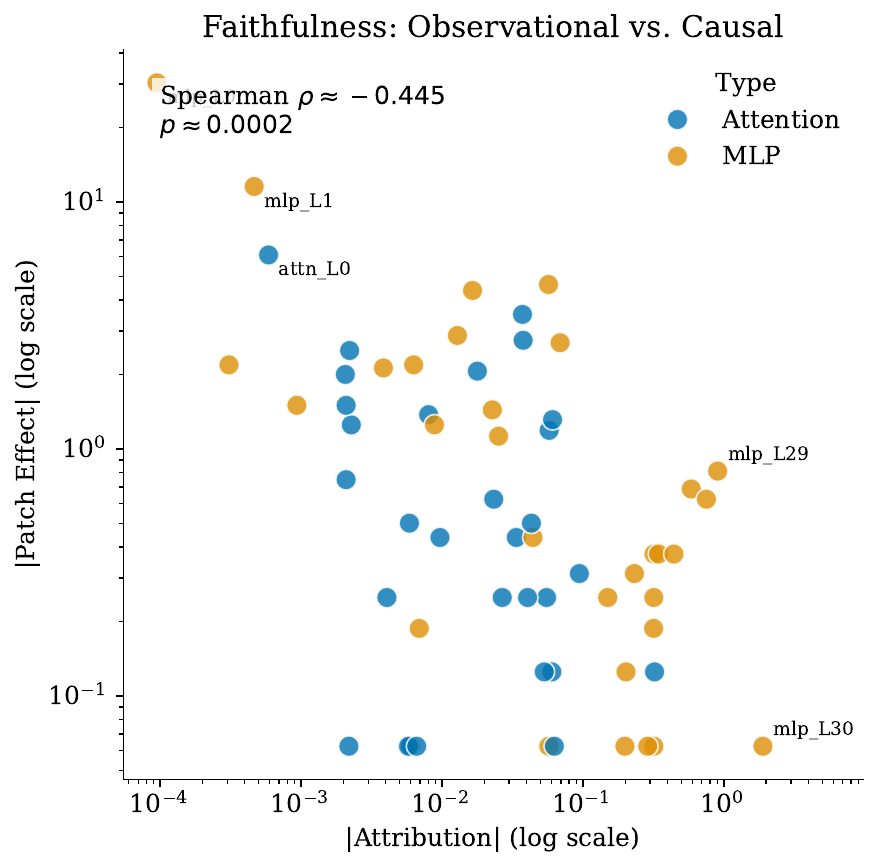}
  \caption{Attribution vs.\ patching, Skywork helpfulness. Late-MLP
  components dominate attribution; early-layer components dominate
  patching.}
  \label{fig:faith}
\end{figure}

At the component level the contrast is sharper than the correlation
suggests. On Skywork helpfulness, top-3 by attribution are
$\code{mlp\_L31} = 5.60$, $\code{mlp\_L30} = 1.90$, $\code{mlp\_L29} = 0.91$
(reward units); top-3 by patching are $\code{mlp\_L0} = 30.31$,
$\code{mlp\_L1} = 11.53$, $\code{attn\_L0} = 6.09$. Late MLPs that
dominate attribution have near-zero patching effects; early-layer
components that dominate patching have unremarkable attribution.
Skywork correctness shows the same pattern with patching led by middle
attention heads (\code{attn\_L11} = 15.75, \code{attn\_L9} = 13.94,
\code{attn\_L14} = 13.03); ArmoRM does too at $\sim$300$\times$ smaller
absolute scales.

\paragraph{Why this is not a bug.} The residual stream is redundant: a
late MLP can be ablated because parallel paths compensate, while early
layers are load-bearing in a way nothing downstream can recover.
Attribution measures contribution to $w_r^\top h_\mathrm{final}$;
patching measures necessity for it to remain high. Equation~(3) does
not promise the two agree. The Pearson correlations sometimes also
flip sign relative to the Spearman ones (ArmoRM correctness Pearson
$r = +0.42$, $p = 0.0005$, but Spearman $\rho = +0.125$, $p = 0.33$),
reflecting sensitivity to single extreme components on a 64-point
sample.

\paragraph{Operational consequence.} Attribution is for exploration;
it is not a substitute for patching when a causal claim is needed.
The framework keeps both views first-class and trivially comparable on
the same components.

\subsection{Cross-model agreement and circuit specialisation}

Reward Lens trajectories of the two architecturally different models
are highly correlated when matched on normalised depth: per-dimension
Pearson correlations of the differential curve are $r = 0.846$
(helpfulness), $0.807$ (safety), $0.844$ (correctness) on a
representative pair. We report this as motivation for the scale study
proposed in \S\ref{sec:proposed}, not as a general result.

Circuit overlap tells a complementary story. The top-$k$ Jaccard
overlap ($k = 10$) between most-attributed components across the four
dimensions is uniform on Skywork (off-diagonal 0.82--1.00, mean per
dimension 0.879) and much more varied on ArmoRM (off-diagonal
0.43--0.67, mean 0.50--0.58). The \emph{trajectory} of preference
formation is similar across architectures, but the \emph{set of
components} carrying that formation is dimension-uniform in a
single-scalar model and dimension-specific in a multi-objective one.

\begin{figure}[H]
  \centering
  \includegraphics[width=0.95\linewidth]{figures/fig3_overlap_Skywork-Reward-Llama-3.1-8B-v0.2.pdf}
  \caption{Cross-model trajectory overlay. Normalised preference differential as a
  function of fractional layer depth for Skywork and ArmoRM on three dimensions.}
  \label{fig:overlay}
\end{figure}

\subsection{Hacking profiles}

The hacking detector runs A/B probes for length, confidence,
formatting, sycophancy, and repetition biases.

\begin{table}[h]
\centering
\begin{tabular}{lrrl}
\toprule
Dimension   & Skywork $d$ & ArmoRM $d$ & Pairs  \\
\midrule
length      & $-1.13$        & $-0.02$       & $n = 3$ \\
confidence  & $-2.08$        & $+2.40$       & $n = 2$ \\
formatting  & $-0.95$        & $+0.91$       & $n = 2$ \\
sycophancy  & $-5.82$        & $-11.92$      & $n = 2$ \\
repetition  & $\infty$ (artefact) & $\infty$ (artefact) & $n = 1$ \\
\bottomrule
\end{tabular}
\caption{Hacking-detector effect sizes (Cohen's $d$). Positive values
mean the model rewards the biased variant. Pair counts are the shipped
defaults and are deliberately small; the repetition entries are
zero-variance and report $\infty$ as an artefact.}
\label{tab:hacking}
\end{table}

The cleanest contrast is the sign flip on confidence and formatting:
Skywork \emph{penalises} both, ArmoRM \emph{rewards} both. Both
strongly penalise sycophancy. Skywork's negative bias on length,
confidence, and sycophancy --- the opposite of the classical
reward-hacking pattern --- is plausibly a signature of its training
recipe. Effect sizes are directional given the small samples; rigorous
claims require larger custom probe sets (see \S\ref{sec:limitations}).

\begin{figure}[H]
  \centering
  \includegraphics[width=0.7\linewidth]{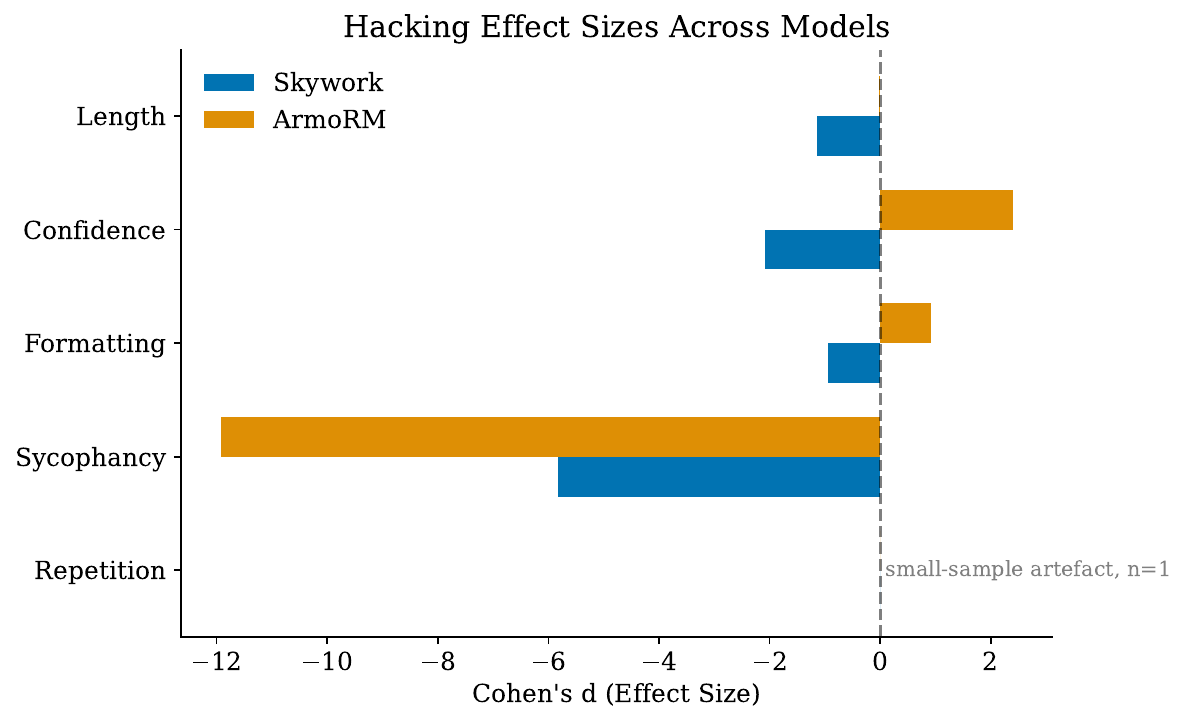}
  \caption{Hacking-detector effect sizes. Positive $d$ rewards the
  biased variant; negative $d$ penalises it.}
  \label{fig:hacking}
\end{figure}

\subsection{Concept-level structure}

We extract six linear concept directions from contrastive preference
pairs --- confidence, formality, agreement, verbosity, hedging, and
helpfulness --- and report each concept's cosine alignment with $w_r$,
the slope of its causal dose-response curve under residual-stream
addition $h \leftarrow h + \alpha v$ for
$\alpha \in \{-2, -1, -0.5, 0.5, 1, 2\}$, and a hacking-risk score
that combines alignment and effect magnitude.

\begin{table}[h]
\centering
\begin{tabular}{lrrrr}
\toprule
& \multicolumn{2}{c}{Skywork} & \multicolumn{2}{c}{ArmoRM} \\
\cmidrule(lr){2-3}\cmidrule(lr){4-5}
Concept     & alignment & causal slope & alignment & causal slope \\
\midrule
agreement   & $0.340$ & $+1.002$  & $0.074$  & $+0.0034$ \\
verbosity   & $0.323$ & $+0.911$  & $-0.052$ & $-0.0032$ \\
helpfulness & $0.290$ & $+0.847$  & $0.080$  & $+0.0079$ \\
formality   & $0.287$ & $+0.825$  & $0.047$  & $+0.0040$ \\
confidence  & $0.261$ & $+0.766$  & $0.062$  & $+0.0063$ \\
hedging     & $0.245$ & $+0.711$  & $0.080$  & $+0.0063$ \\
\bottomrule
\end{tabular}
\caption{Concept reward alignment (cosine of concept direction with
$w_r$) and causal slope (regression of $\Delta$ reward on $\alpha$
over the dose-response curve), per model.}
\label{tab:concepts}
\end{table}

Two findings. First, on Skywork the dose-response curves are clean
straight lines: the agreement concept has alignment $0.340$ and slope
$+1.002$, with negligible deviation from linearity over
$\alpha \in [-2, 2]$ --- a strong signal that the concept is locally
represented as a linear direction. Second, the absolute concept
alignments on ArmoRM are an order of magnitude smaller (max $0.080$
vs.\ $0.340$) and the dose-response slopes correspondingly tiny:
ArmoRM's reward signal does not align cleanly with any single linear
concept direction.

\begin{figure}[H]
  \centering
  \includegraphics[width=0.95\linewidth]{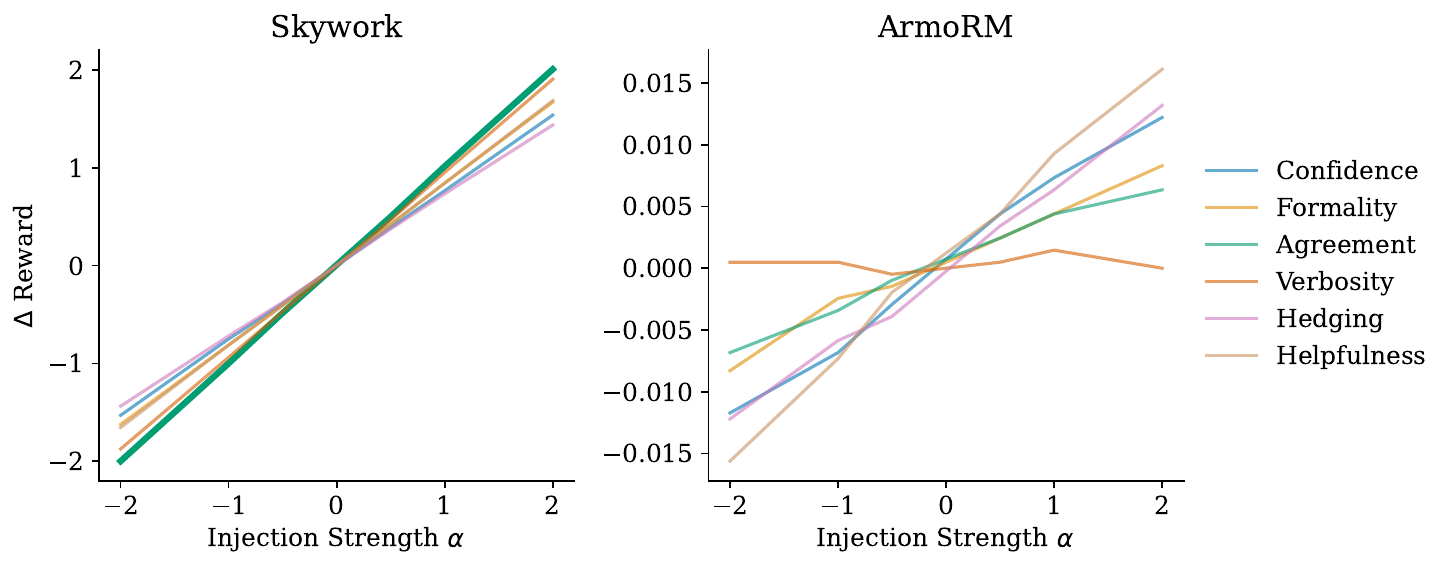}
  \caption{Concept dose-response curves. Skywork responds linearly at
  unit scale; ArmoRM responds at $\sim$1\% the magnitude.}
  \label{fig:concepts}
\end{figure}

\subsection{Reward-term geometry and cascade signal}

The conflict analyser computes cosine similarities between the four
quality-dimension direction vectors. Skywork's matrix is uniformly
aligned (off-diagonal cosines 0.534--0.917); ArmoRM's is aligned but
weaker (0.341--0.954). Neither model has a reward-shape configuration
classified as in-conflict or orthogonal.

The cascade detector reports systemic-risk scores of $1.00$ (Skywork)
and $0.60$ (ArmoRM). \emph{Read this with the small-sample caveat:}
the detector estimates pairwise correlations between
misalignment-dimension scores using the same 1--3 probes per dimension
as the hacking detector; correlations of 2--3-point sequences round to
$\pm 1$ for any non-trivial pair, and the matrix is therefore
degenerate. The cascade-risk score is directional triage, not a
confidence-bearing measurement. The contrast between the two models is
large enough that the ordering may survive a larger sample, but we do
not place weight on the absolute values.

\subsection{Runtime characterisation}

On a single A100-80GB for Skywork (8B, 32 layers): Reward Lens per
pair $\sim$5 s; component attribution per pair $\sim$10 s; full
noising patch over all 64 sublayers $\sim$30 min; full hacking scan
$\sim$5 min; SAE training on a single layer $\sim$8--24 h. Cost is
dominated by the patching pass.

\vspace{1em}
\hrule
\vspace{1em}

\section{Summary of findings}

(1) On Skywork, preferences crystallise uniformly late (fractional
depth $\sim$0.90 across all four dimensions tested); on ArmoRM they
crystallise earlier and with much higher across-pair variance,
consistent with that model's multi-objective architecture.
(2) Late MLPs dominate observational attribution near-universally on
both models, with \code{mlp\_L31} in the top-10 for $\geq 192$ of 200
pairs everywhere we look.
(3) Linear attribution does not predict causal patching effects in
either model (mean Spearman $\rho = -0.256$ on Skywork, $-0.027$ on
ArmoRM); the two quantities should be compared, not substituted.
(4) Cross-model formation-correlation is high on a representative-pair
basis ($r = 0.81$--$0.85$), but circuit-overlap structure differs
sharply: Skywork uses uniform circuits across dimensions, ArmoRM uses
dimension-specific ones.
(5) Skywork and ArmoRM have qualitatively different hacking profiles
--- Skywork penalises confidence and formatting; ArmoRM rewards them;
both strongly penalise sycophancy.
(6) Concept directions extracted from contrastive pairs produce clean
linear dose-response curves on $w_r$ in Skywork (slope $\sim$1 reward
unit per unit $\alpha$ for the agreement concept), at $\sim$1\% the
magnitude on ArmoRM.

\vspace{1em}
\hrule
\vspace{1em}

\section{Proposed experiments and forward-looking work}
\label{sec:proposed}

The framework supports several experiments we have not yet run; we
list them as research questions, not as results, and use them to
motivate two longer-horizon programmes the framework now makes
tractable.

\paragraph{SAE feature-level reward decomposition.} The framework
predicts that for a TopK SAE trained on a chosen residual layer,
$w_r^\top d_i$ summarises each feature's reward contribution. Two
questions follow: is the distribution of $w_r^\top d_i$ heavy-tailed,
indicating a small set of reward-relevant features; and does the
cumulative top-$k$ feature contribution saturate the held-out reward
variance well below $k = n_\mathrm{features}$? We have not run this
and report it as the next high-priority experiment.

\paragraph{Distortion-index validation.} Can the distortion index
predict, \emph{before} any RL, which quality dimension a reward model
will under-cover and therefore allow a downstream policy to hack? The
proposed protocol is to construct a reward-model training run with
one quality dimension deliberately absent from evaluation, score the
resulting model with the distortion analyser, then PPO-train against
it and check whether the predicted dimension is the one that
degrades. This is the boldest test the library affords; success would
move reward-model interpretability from detection to prediction.

\paragraph{Scale study.} How do crystallisation depth,
attribution-vs.-patching faithfulness, and concept alignment vary
along the Skywork-Reward-Gemma-2 ladder (2B / 8B / 27B) and into
Nemotron-4-340B-Reward? Stated, not claimed: we expect crystallisation
to shift later with scale, the faithfulness gap to widen as redundant
pathways multiply, and concept alignment to strengthen with capacity.

\paragraph{Training-loop integration.} The Reward Lens, hacking
detector, and distortion index each have a per-pair runtime small
relative to a reward-model training step on the same hardware. The
natural next step is to turn them into in-loop monitors: a callback
that runs a fixed diagnostic probe set every $N$ optimiser steps and
raises an alarm when distortion exceeds a threshold or a previously
penalised hacking dimension flips sign. This would shift reward-model
interpretability from a post-hoc audit into a training-time
safeguard, before failure modes are baked in. The infrastructure for
the audit half already exists in the library; the integration with
TRL- or OpenRLHF-style trainers is a research programme, not a
feature.

\paragraph{Interpretability-guided reward-model editing.} Once a
concept is identified as both well-aligned with $w_r$ \emph{and}
known-to-be-hackable (\S5.7 already produces such a list on Skywork
--- the agreement concept, slope $+1.002$, is a candidate), the
natural next step is to edit the reward head to suppress that
concept's contribution: subtract its component from $w_r$, retrain
only the head, or apply a low-rank correction. The open question is
whether such edits transfer to a downstream RL run --- whether a
reward model that no longer rewards verbosity in a probe also fails to
reward it in a policy fine-tuned against it. The library's
attribution, patching, SAE, and concept-vector primitives are exactly
the tools needed to localise the target features; what remains is the
edit-and-retrain experiment.

A per-experiment checklist accompanies the source release.

\vspace{1em}
\hrule
\vspace{1em}

\section{Discussion and limitations}
\label{sec:limitations}

\paragraph{Linear decomposition is not causal decomposition.} The
attribution-vs.-patching negative correlation in
\S\ref{sec:faithfulness} is a structural property of linear
decomposition in a redundant residual-stream architecture, not a bug.
The library's own headline empirical result \emph{is} this limitation.
We chose to foreground it in the abstract rather than bury it; an
honest interpretability library should make the gap between
observation and causation legible rather than collapse them.

\paragraph{Small-sample statistics.} The shipped hacking detector uses
1--3 pairs per dimension; Cohen's $d$ on such samples is unstable, the
infinities in Table~\ref{tab:hacking} are zero-variance artefacts, and
the cascade detector's $\pm 1$ pairwise correlations are degenerate.
The fix is straightforward: scale the diagnostic preference set by
10--30$\times$ and add bootstrap confidence intervals at the library
level.

\paragraph{Single-pair faithfulness.} The faithfulness experiment of
\S\ref{sec:faithfulness} is per-dimension single-pair; it tests a
specific example rather than the population. The headline conclusion
(attribution is not faithful to patching) is robust enough across six
dimension/model combinations to be reported, but per-dimension numbers
should be interpreted as point estimates. The cross-model formation
correlations in \S5.5 share this single-pair caveat.

\paragraph{Observational caveats throughout.} The Reward Lens,
component attribution, concept alignments, and SAE feature alignments
are all observational. Only activation patching (and divergence-aware
patching) provides causal evidence. The library says this; this paper
says this; downstream work should say this.

\paragraph{Inherited open problems.} \code{reward-lens} does not solve
the known open problems of residual-stream interpretability: SAEs
remain lossy; activation patching can induce divergent representations
(the divergence-aware patcher is a partial mitigation, not a
solution); cross-layer circuit discovery is underdeveloped. What the
library does is make these problems testable in a new model class ---
and in some cases the scalar-output structure may let us study them
with less clutter than in generative models.

\paragraph{Reward models we cannot analyse.} Proprietary API-only
reward models inside frontier-lab training pipelines are closed to any
weight-level interpretability tool, this one included. Meaningful
mechanistic science on reward models requires open-weight reward
models; the recent uptick in open releases is encouraging.

\paragraph{Process reward models.} Per-step scoring works, but
step-boundary detection, accumulated-quality tracking, and step-level
attribution are not implemented in the current release and are a known
gap.

\vspace{1em}
\hrule
\vspace{1em}

\section{Conclusion}

Reward models are the object that defines what an RLHF-trained
language model is optimising for, and they are the least mechanistically
understood component of the alignment pipeline. \code{reward-lens}
packages a Reward Lens, component attribution, contrastive activation
patching, hacking probes, concept-vector analysis, and SAE-feature
attribution against a single organising principle --- the reward
head's weight vector $w_r$ as the axis of all decomposition --- behind
one \code{pip install}. The headline empirical result is negative and
honest: on two production reward models, linear attribution does not
predict causal importance, and researchers should plan their
experiments accordingly. The library was built so that this gap could
be measured rather than ignored; the natural next steps it enables ---
in-training distortion monitoring and interpretability-guided
reward-head editing --- treat reward-model interpretability as
something that should run alongside reward-model training, not after
it.

\vspace{1em}
\hrule
\vspace{1em}

\section*{References}

\begin{itemize}\itemsep0pt
\item Belrose, N., Furman, Z., Smith, L., et al. (2023). Eliciting
  Latent Predictions from Transformers with the Tuned Lens.
  \emph{arXiv:2303.08112}.
\item Bloom, J. (2024). SAELens (software).
\item Christiano, P., Leike, J., Brown, T., Martic, M., Legg, S.,
  Amodei, D. (2017). Deep Reinforcement Learning from Human
  Preferences. \emph{NeurIPS}.
\item Conmy, A., Mavor-Parker, A., Lynch, A., Heimersheim, S.,
  Garriga-Alonso, A. (2023). Towards Automated Circuit Discovery for
  Mechanistic Interpretability. \emph{NeurIPS}.
\item Fiotto-Kaufman, J., et al. (2024). NNsight and NDIF:
  Democratizing Access to Foundation Model Internals. \emph{arXiv}.
\item Gao, L., et al. (2024). Scaling and Evaluating Sparse
  Autoencoders. \emph{OpenAI}.
\item Lambert, N., et al. (2024). RewardBench: Evaluating Reward
  Models for Language Modeling. \emph{arXiv:2403.13787}.
\item Liu, C. Y., et al. (2024). Skywork-Reward: Bag of Tricks for
  Reward Modeling in LLMs.
\item Nanda, N., Bloom, J. (2022). TransformerLens (software).
\item nostalgebraist (2020). Interpreting GPT: the logit lens.
  \emph{LessWrong}.
\item Ouyang, L., et al. (2022). Training language models to follow
  instructions with human feedback. \emph{NeurIPS}.
\item Park, R., et al. (2024). Disentangling Length from Quality in
  Direct Preference Optimization. \emph{arXiv:2403.19159}.
\item Perez, E., et al. (2023). Discovering Language Model Behaviors
  with Model-Written Evaluations. \emph{ACL}.
\item Rajamanoharan, S., et al. (2024). Gemma Scope.
\item Sharma, M., et al. (2024). Towards Understanding Sycophancy in
  Language Models. \emph{ICLR}.
\item Singhal, P., et al. (2024). A Long Way to Go: Investigating
  Length Correlations in RLHF. \emph{arXiv:2310.03716}.
\item Stiennon, N., et al. (2020). Learning to summarize with human
  feedback. \emph{NeurIPS}.
\item Syed, A., Nanda, N. (2024). Attribution Patching Outperforms
  Automated Circuit Discovery.
\item Wang, H., et al. (2024). ArmoRM: Interpretable Preferences via
  Multi-Objective Reward Modeling with Mixture-of-Experts.
  \emph{arXiv:2406.12845}.
\item Chen, R., et al. (2025). Persona Vectors: Monitoring and
  Controlling Character Traits in Language Models. Anthropic.
  \emph{arXiv:2507.21509}.
\item Addressing Divergent Representations from Causal Interventions
  on Neural Networks (2025). \emph{arXiv:2511.04638}.
\item MacDiarmid, M., Wright, B., Uesato, J., Benton, J., Kutasov, J.,
  Price, S., et al. (2025). Natural Emergent Misalignment from Reward
  Hacking in Production RL. Anthropic. \emph{arXiv:2511.18397}.
\item Wang, J., \& Huang, J. (2026). Reward Hacking as Equilibrium
  under Finite Evaluation. \emph{arXiv:2603.28063}.
\item Kaufmann, M., Lindner, D., Zimmermann, R. S., \& Shah, R.
  (2026). Aligned, Orthogonal or In-conflict: When can we safely
  optimize Chain-of-Thought? Google DeepMind.
  \emph{arXiv:2603.30036}.
\end{itemize}

\appendix

\vspace{1em}
\hrule
\vspace{1em}

\section{Library overview and design principles}
\label{app:design}

This appendix is the technical manual for \code{reward-lens}. It is
organised around the conceptual through-line of the paper: how does
mechanistic interpretability change when the head is a single linear
functional $w_r$ rather than a vocabulary unembedding $W_U$? Each
primitive is paired with its generative-LLM antecedent, the change is
identified, and the reward-model implementation is given in enough
detail that a reader could replicate it on a different framework.

\subsection{Why HuggingFace-native rather than TransformerLens-bound}

TransformerLens (Nanda \& Bloom, 2022) is the de-facto substrate of
generative-model mechanistic interpretability. Its primitives ---
\code{HookedTransformer}, the logit lens, direct logit attribution,
attribution patching --- are typed against the assumption that the
output of the network is a vocabulary distribution. Two specific
assumptions break for a reward model. First, the unembedding $W_U$ is
loaded as a known tensor of shape
$\lvert V \rvert \times d_\mathrm{model}$; a reward model has no
unembedding, only a one-row score head whose extraction varies by
family. Second, the lens, attribution, and patching APIs return
quantities indexed by token id, with utility functions for token
selection and target-distribution comparison; for a reward model the
target is a scalar with no token id, and these utilities are dead code
or worse, misleading. A from-scratch port that respects the new shape
of the head is cleaner than monkey-patching.

The cost of this choice is that \code{reward-lens} re-implements the
hook machinery on top of plain HuggingFace forward hooks. The benefit
is coverage of model families with non-standard heads, including
multi-objective and MoE-gated heads, and a single import path
(\code{from reward\_lens import RewardModel}) that wraps anything
loadable through \code{AutoModelForSequenceClassification}.

\subsection{Why an adapter protocol, and where its boundary sits}

Llama, Mistral, Gemma-2, and ArmoRM all expose layers, attention
sublayers, and MLP sublayers under different attribute paths
(\code{model.model.layers}, \code{model.transformer.h},
\code{model.gpt\_neox.layers}, etc.); reward heads are sometimes
\code{score}, sometimes \code{classifier}, sometimes a multi-tensor
object behind a gating network. The adapter protocol isolates these
family-specific details into a ten-method interface and keeps every
other module in the library written against that interface. The
boundary is drawn so that the adapter handles \emph{everything family
specific} (head extraction, layer iteration, sublayer access, output
parsing) and \emph{nothing else}: the lens, attribution, patcher,
hacking detector, SAE, and extension modules are family-agnostic.

\subsection{Why observational and causal views are kept distinct}

A recurring failure mode of interpretability work is to read off a
contribution and report it as if it were a cause. The experiments of
\S\ref{sec:faithfulness} show why that conflation is dangerous in this
setting: on the same components, the two quantities can have negative
rank correlation. The library makes the distinction syntactic: lens
and component attribution return \code{ComponentResult}; the patcher
returns \code{PatchingResult}; both have the same component schema, so
a side-by-side plot is one line. Equation~(3) is taught in
\S\ref{sec:core} as an algebraic identity on contributions, not a
claim about counterfactual effects.

\subsection{Package surface, by purpose}

The user-facing surface clusters into five groups.
\emph{Model wrapping}: \code{RewardModel} (the wrapper) and the
\code{ModelAdapter} hierarchy (\code{LlamaAdapter},
\code{MistralAdapter}, \code{Gemma2Adapter}, \code{ArmoRMAdapter},
\code{GenericAdapter}, with a \code{get\_adapter} factory).
\emph{Observational primitives}: \code{RewardLens}
(\code{RewardLensResult}, \code{reward\_lens\_plot}),
\code{ComponentAttribution} (\code{ComponentResult}), and the
\code{TopKSAE} family (\code{TopKSAE}, \code{ActivationCollector},
\code{SAETrainer}, \code{FeatureAnalyzer}).
\emph{Causal primitives}: \code{ActivationPatcher}
(\code{PatchingResult}) and \code{DivergenceAwarePatching}
(\code{DivergenceAwarePatchingResult}).
\emph{Vulnerability and theory-grounded extensions}:
\code{HackingDetector} (\code{HackingReport}),
\code{DistortionAnalyzer} (\code{DistortionReport}),
\code{MisalignmentCascadeDetector} (\code{CascadeReport}),
\code{RewardConflictAnalyzer} (\code{ConflictReport}),
\code{ConceptExtractor} (\code{ConceptAlignmentReport}, with
\code{quick\_concept\_analysis}).
\emph{Supporting infrastructure}: \code{Comparator} for cross-model
analyses, \code{diagnostic\_data} for curated preference pairs, and
\code{viz} for shared plotting.

\section{The model wrapper in depth}
\label{app:wrapper}

\code{RewardModel} wraps a HuggingFace transformer + score head and
exposes the single object on which all other primitives act.

\paragraph{Reward-head extraction.} The adapter's
\code{get\_reward\_head\_params(model)} returns
$(w_r, b_r)$ with $w_r \in \mathbb{R}^{d_\mathrm{model}}$. For
single-scalar heads (Llama, Mistral, Gemma-2 trained as RMs) the
weight is the single row of the score linear layer. For ArmoRM the
weight is the column-mean of the 19-row \code{regression\_layer}
weight matrix --- the right approximation for aggregate analyses ---
and a separate \code{per\_objective\_directions} method returns the
full $19 \times d_\mathrm{model}$ matrix for dimension-resolved
analyses such as the conflict analyser. For models with truly
nonlinear heads, the protocol is incompatible by design: a primitive
that decomposes along $w_r$ does not have a meaningful target.

\paragraph{What the cache contains.} \code{forward\_with\_cache}
returns an \code{ActivationCache} dataclass with: a dict of per-layer
residual streams indexed $-1, 0, \ldots, L-1$ (where $-1$ is the
post-embedding output, $\ell$ is the post-block residual at layer
$\ell$), a dict of per-layer attention sublayer outputs, a dict of
per-layer MLP sublayer outputs, and the final-token positions used to
index out the relevant activations. Each cached tensor is the
final-token slice
$h^{(\ell)}[\mathrm{batch}, \mathrm{seq\_lengths} - 1, :]$ unless
\code{cache\_full\_sequences=True}, in which case the full
$(\mathrm{batch}, \mathrm{seq}, d_\mathrm{model})$ tensor is also
retained. Patching needs the full tensors to splice along the shared
prefix; the lens and attribution need only the final-token slice.

\paragraph{Hook lifecycle.} Hooks are registered as a context manager
via \code{rm.hooks()}. The wrapper is reentrant: a forward pass
inside a hook block sees the registered hooks; a forward pass outside
does not. On exit, all handles are removed even if the body raised. A
single \code{forward\_with\_cache} call sets up its own hooks
internally; user code rarely needs the lower-level interface.

\paragraph{Layer-norm handling.} Decoder-only transformers apply a
final layer norm before the score head:
$r = w_r^\top \mathrm{LN}(h_\mathrm{final}) + b_r$. We absorb this
norm conceptually into $w_r$ for analytic purposes (the lens reports
$w_r^\top h^{(\ell)}$, not $w_r^\top \mathrm{LN}(h^{(\ell)})$), but
the forward pass keeps the norm in place when computing actual
rewards; the small inconsistency between intermediate and final
projection is the price of a tractable linear story. In practice the
crystallisation depths reported in \S5.2 are insensitive to whether
the norm is applied at every layer or only at the output.

\paragraph{Final-token selection.} Sequences are batched with
right-padding; the wrapper uses
$\mathrm{final\_pos} = (\mathrm{attention\_mask}.\mathrm{sum}(1) - 1).\mathrm{clamp}(0, T-1)$
to recover per-sample final-token indices. The clamp is defensive
against zero-length inputs.

\paragraph{Device, dtype, and inference-mode discipline.} The wrapper
runs the underlying model in \code{torch.inference\_mode()} and calls
\code{rm.eval()}; cached activations are detached and moved to CPU
once captured to avoid GPU-memory growth across a long run; user
tensors stay on the model's device. Bfloat16 is supported throughout;
arithmetic that requires float32 (the SAE loss, conflict cosine
similarities) is up-cast locally.

\section{The adapter protocol in depth}
\label{app:adapter}

The adapter abstraction is a ten-method interface (Table~\ref{tab:adapter-methods}).
The shipped adapters are summarised in
Table~\ref{tab:adapter-families}; ArmoRM is then expanded in its own
subsection.

\begin{table}[h]
\centering
\small
\begin{tabular}{lll}
\toprule
Method & Returns & Contract \\
\midrule
\code{get\_reward\_head\_params(model)} & $(w_r, b_r)$ & $w_r \in \mathbb{R}^{d_\mathrm{model}}$, $b_r \in \mathbb{R}$ \\
\code{get\_layers(model)} & \code{nn.ModuleList} & ordered transformer blocks \\
\code{n\_layers(model)} & \code{int} & number of blocks \\
\code{n\_heads(model)} & \code{int} & attention heads per block \\
\code{get\_attn\_module(layer)} & \code{Optional[nn.Module]} & per-block attention sublayer \\
\code{get\_mlp\_module(layer)} & \code{Optional[nn.Module]} & per-block MLP sublayer \\
\code{extract\_layer\_output(out)} & \code{Tensor[B,T,d]} & residual after the block \\
\code{extract\_attn\_output(out)} & \code{Tensor[B,T,d]} & attention residual contribution \\
\code{extract\_mlp\_output(out)} & \code{Tensor[B,T,d]} & MLP residual contribution \\
\code{get\_embedding(model)} & \code{nn.Module} & input embedding layer \\
\code{extract\_reward(out, inputs)} & \code{Tensor[scalar]} & final scalar reward \\
\bottomrule
\end{tabular}
\caption{The adapter contract. Implementing these eleven entries
(plus the standard \code{\_\_init\_\_}) is sufficient to bring up a
new model family.}
\label{tab:adapter-methods}
\end{table}

\begin{table}[h]
\centering
\small
\begin{tabular}{lll}
\toprule
Adapter & Layer path & Reward head \\
\midrule
\code{LlamaAdapter}   & \code{model.model.layers}   & \code{model.score} (single-scalar) \\
\code{MistralAdapter} & \code{model.model.layers}   & \code{model.score} (single-scalar) \\
\code{Gemma2Adapter}  & \code{model.model.layers}   & \code{model.score} (single-scalar) \\
\code{ArmoRMAdapter}  & \code{model.model.layers}   & 19-objective regression + MoE gate \\
\code{GenericAdapter} & auto-detected               & auto-detected \\
\bottomrule
\end{tabular}
\caption{Shipped adapters. \code{GenericAdapter} walks the module tree
for layer paths matching
\code{[model$\,|\,$transformer$\,|\,$gpt\_neox].[layers$\,|\,$h]} and
reward heads named
\code{score / classifier / v\_head / reward\_head}, falling back to
the last \code{nn.Linear} in an \code{nn.Sequential}.}
\label{tab:adapter-families}
\end{table}

\subsection{ArmoRM as the stress test}

ArmoRM-Llama3-8B-v0.1 has three components fused on top of the
backbone:
(i) a \code{regression\_layer} of shape
$19 \times d_\mathrm{model}$ that maps the final hidden state to 19
per-objective scores;
(ii) a prompt-conditional MoE gating network that, given the prompt's
final hidden state, outputs 19 mixture weights summing to one; and
(iii) a debiasing \code{reward\_transform\_matrix} applied to the
gated combination to produce the final scalar reward
\code{output.score}. The adapter's
\code{get\_reward\_head\_params} returns the column-mean
$\bar{w} = \tfrac{1}{19}\sum_{k=1}^{19} W_{\mathrm{reg},k}$, which is
the right approximation for analyses that aggregate across objectives
(the lens reads off ``average reward direction''). For
dimension-resolved analyses the adapter exposes
\code{per\_objective\_directions}, returning the full
$19 \times d_\mathrm{model}$ matrix; the conflict analyser is then
called directly on those rows. We deliberately do not try to absorb
the MoE gate into a single direction: the gate is prompt-conditional,
so the effective direction varies across inputs, and any single
``direction'' that pretends otherwise would be misleading. Aggregate
analyses use $\bar{w}$ and accept the approximation; per-objective
analyses use the full matrix and respect the true geometry.

\subsection{Adding a new family}

For a model whose backbone matches the auto-detection patterns of
\code{GenericAdapter}, no code change is required: the wrapper picks
up the right layers and head on first call. For a model with a
non-standard layer path or head shape, an adapter implementing the
eleven-method interface is typically about fifty lines.

\section{Core primitives, in depth}
\label{app:core}

This appendix is structured one subsection per primitive. Each opens
with the generative-LLM antecedent, identifies what changes for a
reward model, and gives the reward-model implementation in enough
detail to be reproduced. Limitations specific to the primitive (as
opposed to the global limitations in \S\ref{sec:limitations}) are
called out in line.

\subsection{Reward Lens}

\paragraph{Generative antecedent.} The logit lens (nostalgebraist,
2020) takes the residual stream at layer $\ell$ and applies the
unembedding to obtain a vocabulary distribution
$p^{(\ell)} = \mathrm{softmax}(W_U \mathrm{LN}(h^{(\ell)}))$, then
reads off the lens's prediction for the next token. Tuned-lens
(Belrose et al., 2023) replaces $W_U$ with an affine transform
trained to match the final-layer prediction, sharpening early-layer
readouts.

\paragraph{What changes for reward models.} The vocabulary projection
is replaced by the reward direction
$r^{(\ell)} = w_r^\top h^{(\ell)} + b_r$; the target is a scalar, not a
distribution; there is no per-token bookkeeping. The natural readout
on a preference pair is the per-layer differential
$\Delta^{(\ell)} = r^{(\ell)}_{\mathrm{pref}} - r^{(\ell)}_{\mathrm{dispref}}$,
which inherits the contrastive structure of the data without extra
machinery.

\paragraph{Implementation.} \code{RewardLens.trace(prompt, pref, dispref)}
runs two cached forwards (one per completion), pulls out
$\{h^{(\ell)}\}_{\ell=-1}^{L-1}$ at the final token, and returns a
\code{RewardLensResult} with fields \code{layers},
\code{reward\_lens\_preferred},
\code{reward\_lens\_dispreferred},
\code{differential},
\code{marginal\_contributions} (per-layer first differences of
\code{differential}), and the final scalars \code{reward\_preferred},
\code{reward\_dispreferred}. The \emph{crystallisation layer} is
defined as the smallest $\ell$ for which $\Delta^{(\ell)}$ reaches
50\,\% of $\Delta^{(L-1)}$ in the same sign; for a $\sim$0 final
differential the crystallisation depth is undefined and the field is
returned as \code{None}.

\paragraph{Plotting and aggregation.} \code{RewardLensResult.plot}
draws preferred/dispreferred trajectories with the crystallisation
layer marked. \code{reward\_lens\_plot(model, prompt, pref, dispref)}
is the one-call convenience. Aggregation across a dataset is plain
Python over a list of results --- the dataclass is intentionally
flat.

\paragraph{Primitive-specific limitation.} The lens reports
$w_r^\top h^{(\ell)}$ without applying the final layer norm at
intermediate layers; intermediate-layer projections are therefore on a
slightly different scale than the final layer. This affects the
absolute magnitudes (which we never compare across layers) but not the
crystallisation index.

\subsection{Component attribution}

\paragraph{Generative antecedent.} Direct logit attribution decomposes
$h_\mathrm{final} = h_\mathrm{embed} + \sum_{\ell}(\mathrm{attn}^{(\ell)} + \mathrm{mlp}^{(\ell)})$
and projects each summand through $W_U$ to get a token-indexed
contribution per component, summarised against the difference of two
target logits.

\paragraph{What changes.} The projection is $w_r^\top$, not $W_U$;
the target is the scalar reward differential, not a logit difference.
The output is a flat vector of $2L+1$ signed scalars (plus the bias)
rather than a $(2L+1) \times \lvert V \rvert$ matrix. The bookkeeping
is correspondingly simpler.

\paragraph{Implementation.}
\code{ComponentAttribution.attribute(prompt, pref, dispref)} caches
both forwards, computes
$c_\mathrm{embed} = w_r^\top h_\mathrm{embed}$,
$c^{(\ell)}_{\mathrm{attn}} = w_r^\top \mathrm{attn}^{(\ell)}$, and
$c^{(\ell)}_{\mathrm{mlp}} = w_r^\top \mathrm{mlp}^{(\ell)}$ for both
completions, and returns a \code{ComponentResult} with fields
\code{component\_names} (e.g.\ \code{embed}, \code{attn\_L0},
\code{mlp\_L0}, \ldots), \code{component\_types}, \code{layer\_indices},
\code{contributions\_preferred}, \code{contributions\_dispreferred},
\code{differential\_contributions}, and the totals
\code{total\_reward\_preferred}, \code{total\_reward\_dispreferred}.
Utilities \code{top\_k(k, by="differential"|"preferred"|"dispreferred")},
\code{by\_type("attn"|"mlp"|"embed")}, \code{plot\_top\_k},
\code{plot\_heatmap} act on this dataclass.

\paragraph{Sanity check.} The sum of all \code{contributions} plus
$b_r$ should match the model's own reward to floating-point
precision; this is asserted in the wrapper's tests and is the cleanest
correctness check the library has.

\paragraph{Primitive-specific limitation.} Attribution is purely
observational; the gap between attribution and patching is the
content of \S\ref{sec:faithfulness}.

\subsection{Activation patching}

\paragraph{Generative antecedent.} Activation patching (Heimersheim et
al.; Syed \& Nanda, 2024) replaces a component's output on a clean
input with the corresponding output from a corrupted input (or vice
versa, or with zeros) and measures the change in a target logit
difference.

\paragraph{What changes.} The contrastive structure of preference data
gives the patcher its (clean, corrupted) pair for free: noising
patching takes the preferred completion as clean and the dispreferred
as corrupted; denoising swaps. The readout is
$\Delta r = w_r^\top (h_\mathrm{final}^{\mathrm{clean}} - h_\mathrm{final}^{\mathrm{patched}})$,
a single scalar per patched component. The same component schema as
component attribution allows trivial side-by-side comparison
(\S\ref{sec:faithfulness}).

\paragraph{Modes.} Three modes are supported.
Table~\ref{tab:patching-modes} summarises their semantics and when
each is the right choice.

\begin{table}[h]
\centering
\small
\begin{tabular}{llll}
\toprule
Mode & Source $\to$ target & Measures & Use when \\
\midrule
noising    & dispref $\to$ pref     & necessity & you want ``which components must be intact''\\
denoising  & pref $\to$ dispref     & sufficiency & you want ``which components alone flip the call''\\
zero ablation & zeros $\to$ pref    & component-removal effect & no contrastive pair available \\
\bottomrule
\end{tabular}
\caption{Patching modes. Noising is the default; denoising is the
symmetric choice; zero ablation is OOD by construction and should be
read with the divergence-aware patcher's results in mind.}
\label{tab:patching-modes}
\end{table}

\paragraph{Sequence-length handling.} Source and target sequences are
typically of different lengths. The hook truncates the source
activation to the target's length and copies along the shared prefix:
$h_\mathrm{patched}[:, :T_\mathrm{shared}, :] = h_\mathrm{source}[:, :T_\mathrm{shared}, :]$,
$h_\mathrm{patched}[:, T_\mathrm{shared}:, :] = h_\mathrm{target}[:, T_\mathrm{shared}:, :]$.
For preference pairs that share a prompt and differ only in the
completion, this is the right thing. Path-patching at the
token-position level is future work.

\paragraph{Implementation.}
\code{ActivationPatcher.patch\_all\_components(prompt, pref, dispref, mode)}
iterates over all $2L$ sublayers (and optionally the embedding),
re-running the model with the splice hook installed for one component
at a time, and returns a \code{PatchingResult} with the same
component schema as \code{ComponentResult} plus
\code{patch\_effects} (signed, in reward units) and
\code{original\_differential}. \code{normalized\_effects} divides by
\code{original\_differential} so that 1.0 means ``patching this
component fully kills the preference''.

\paragraph{Primitive-specific limitations.} Patching can induce
out-of-distribution activations, especially zero ablation. The
divergence-aware patcher (\S\ref{app:ext-divergence}) is the
diagnostic; it does not eliminate the problem.

\subsection{Hacking detector}

\paragraph{Generative antecedent.} Behavioural red-teaming of language
models, as in Perez et al.\ (2023), is end-to-end: probe inputs are
fed to the policy and outputs are scored. The reward model is invisible
to the probe.

\paragraph{What changes.} Inside the library the probe is fed
\emph{directly} to the reward model and the scalar output is the
readout, with no intermediate generation. A probe is an A/B pair of
completions identical except for the dimension being tested.

\paragraph{Implementation.} \code{HackingDetector(model).scan(tests=...)}
runs five default dimensions (Table~\ref{tab:hacking-dims}); each
dimension's pairs are scored to produce a vector of deltas
$\delta_i = r(\mathrm{biased}) - r(\mathrm{neutral})$. Cohen's $d$ is
$\bar\delta / \mathrm{std}(\delta)$. The result is a
\code{HackingReport} with one \code{BiasTestResult} per dimension
(fields: \code{dimension}, \code{reward\_deltas}, \code{mean\_delta},
\code{std\_delta}, \code{effect\_size}, \code{pairs\_tested},
\code{verdict}). Custom probe sets are supplied via
\code{test\_custom\_pair(prompt, neutral, biased)}.

\begin{table}[h]
\centering
\small
\begin{tabular}{ll}
\toprule
Dimension & A/B contrast \\
\midrule
length      & identical content, $1\times$ vs $3$--$4\times$ length \\
confidence  & hedged uncertainty vs authoritative tone, same content \\
formatting  & plain prose vs markdown headers/bullets, same content \\
sycophancy  & honest correction vs flattering agreement on a false claim \\
repetition  & single statement vs same fact restated three times \\
\bottomrule
\end{tabular}
\caption{Default hacking-detector dimensions.}
\label{tab:hacking-dims}
\end{table}

\paragraph{Primitive-specific limitations.} The shipped probe sets are
diagnostic, not statistical. Reliable hacking claims require
user-supplied probe sets with $n \geq 30$ pairs per dimension and
bootstrap confidence intervals; the small zero-variance entries in
\S5.6 are an honest artefact of the shipped defaults.

\subsection{TopK sparse-autoencoder feature attribution}

\paragraph{Generative antecedent.} SAE-based circuit discovery (Bloom,
2024; Gao et al., 2024; Rajamanoharan et al., 2024) trains a
dictionary on residual-stream activations of a generative LLM and
analyses each feature by its top-activating tokens or by its
contribution through $W_U$ to a target logit.

\paragraph{What changes.} Equation (5) gives a closed-form per-feature
reward alignment $w_r^\top d_i$ that requires no further forward pass
once the SAE is trained. The notion of ``top-activating token'' is
replaced by ``top-activating preference pair'', or simply by the
per-feature alignment magnitude.

\paragraph{Implementation.} \code{TopKSAE(d\_model, n\_features, k)}
implements the standard TopK SAE: encoder
$z = (x - b_\mathrm{dec})W_\mathrm{enc} + b_\mathrm{enc}$,
$f = \mathrm{ReLU}(\mathrm{TopK}(z, k))$, decoder
$\hat x = f W_\mathrm{dec} + b_\mathrm{dec}$. The training loss is
$\lVert x - \hat x \rVert^2 + 0.01 \lVert \mathrm{cols}(W_\mathrm{dec})\!-\!1 \rVert^2$;
decoder columns are renormalised to unit norm after each step.
\code{ActivationCollector} streams residual-stream activations from
a list of (prompt, response) inputs at a specified layer.
\code{SAETrainer} runs cosine-annealed Adam.
\code{FeatureAnalyzer} returns a \code{FeatureInfo} list with fields
\code{reward\_alignment} ($w_r^\top d_i$),
\code{mean\_activation},
\code{activation\_frequency},
\code{top\_activating\_indices},
\code{top\_activating\_values}; the methods
\code{top\_reward\_features(k)},
\code{bottom\_reward\_features(k)},
\code{decompose\_reward\_for\_input(x, w\_r)} surface the standard
analyses.

\paragraph{Primitive-specific limitations.} SAEs are lossy: the
reconstruction error in equation (5) is non-zero, and a small
fraction of the reward variance is unexplained even by the full
feature set. Training cost ($\sim$8--24 h per layer on a single
A100) is the most expensive item in the library.

\subsection{Cross-model comparator}

\paragraph{Generative antecedent.} Cross-model comparison in the
generative case has typically been ad-hoc: align two models'
intermediate activations by depth fraction and visualise.

\paragraph{What changes.} Because the lens is one-dimensional, the
analogue is short: per-model differential trajectories are scalar
functions of normalised depth, so a Pearson correlation on the
shared depth grid is the natural summary.

\paragraph{Implementation.} \code{Comparator([modelA, modelB,
\ldots]).compare(prompt, pref, dispref)} runs the lens (and optionally
attribution) on each model, normalises the differential to fractional
depth $(\ell + 1)/(L + 1)$, interpolates onto a common grid, and
returns a \code{ComparisonResult} with
\code{lens\_results} (per model),
\code{attribution\_results} (per model, optional),
\code{crystallization\_layers} (per model, fractional),
\code{formation\_correlations} (pairwise Pearson). The cross-model
overlay of \S5.4 is one call.

\section{Theory-grounded extensions, in depth}
\label{app:extensions}

Each extension operationalises a recent alignment-theory result. The
science is due to the cited authors; the contribution of
\code{reward-lens} is the operationalisation in code, the integration
with the rest of the library, and the production of the dataclass-typed
output that the reader can act on.

\subsection{Distortion index}
\label{app:ext-distortion}

\paragraph{Underlying theory.} Reward Hacking as Equilibrium under
Finite Evaluation (Wang \& Huang, 2026) argues that reward hacking is the equilibrium
strategy when the policy can identify quality dimensions left
under-covered by evaluation. The actionable corollary is that one
can predict which dimensions are at risk \emph{before} any RL by
quantifying coverage from the evaluation suite alone.

\paragraph{Operationalisation.} Inputs: a list of quality dimensions
(strings) and a list of evaluation probes, each tagged with the
dimensions it targets. For each probe-dimension pair the analyser
computes a discrimination score
$c_{ij} = (\Delta r_{ij} - \Delta r_\mathrm{min}) /
(\Delta r_\mathrm{max} - \Delta r_\mathrm{min})$
from the reward delta on that probe's preference pair. Per-dimension
effective coverage uses the diminishing-returns aggregation
$C_\mathrm{eff}(d) = 1 - \prod_{i\,:\,d \in \mathrm{tags}(i)}(1 - c_{id})$,
which dominates a sum when several weak probes target a dimension.
Distortion is
$D(d) = 1 - C_\mathrm{eff}(d) / \max_{d'} C_\mathrm{eff}(d')$.

\paragraph{Output.} \code{DistortionReport} with fields
\code{quality\_dimensions},
\code{per\_dimension\_distortion}: $\{d \to D(d)\}$,
\code{under\_covered\_dimensions} (those above the distortion
threshold),
\code{predicted\_hacking\_severity},
\code{coverage\_matrix} ($n_\mathrm{probes} \times n_\mathrm{dims}$),
\code{effective\_coverage} per dimension, \code{recommendations}.

\paragraph{Agentic amplification.}
\code{analyze\_agentic\_amplification(base\_dimensions, tool\_count)}
applies a heuristic amplification factor reflecting that agentic
quality dimensions scale combinatorially with available tools while
evaluation scales linearly: amplified distortion =
base distortion $\times \log_2(2^T/(T+1) + 1)$, capped at 1.

\paragraph{Worked interpretation.} On a probe set that omits an
``honesty'' dimension entirely, $D(\mathrm{honesty}) = 1$ regardless
of the rest, and the report flags honesty as the predicted
under-covered dimension. The natural validation experiment is in
\S\ref{sec:proposed}.

\subsection{Divergence-aware patching}
\label{app:ext-divergence}

\paragraph{Underlying theory.} Addressing Divergent Representations
from Causal Interventions on Neural Networks (2025, arXiv:2511.04638) shows that activation patching can
push activations off the manifold of normal forward passes; a
non-trivial fraction of large patch effects are spurious in the
sense that they activate circuits the model never reaches in
distribution. The fix is to detect off-manifold activations and
classify them.

\paragraph{Operationalisation.} A \code{DistributionEstimator} fits a
per-component Gaussian by collecting clean-forward activations on a
provided set of (prompt, response) inputs and storing means and
covariances. At patch time, the patched activation $a$ is scored
against its component's distribution by Mahalanobis distance
$D = \sqrt{(a-\mu)^\top \Sigma^{-1}(a-\mu)}$. Activations with
$D > 2$ (in $\sigma$ units, the default threshold) are flagged.

\paragraph{Classification.} For each flagged component the patcher
classifies the divergence as
\emph{harmless} (off-manifold but
$|\mathrm{patch\,effect}| < 0.1 \cdot |\mathrm{original\,differential}|$,
plausibly null-space) or
\emph{pernicious} (off-manifold and
$|\mathrm{patch\,effect}| \geq 0.1 \cdot |\mathrm{original\,differential}|$,
plausibly a spurious circuit activation). A reliability score
$1 - n_\mathrm{pernicious}/n_\mathrm{components}$ is returned with the
result; runs with reliability below 0.7 are flagged in the report.

\paragraph{Output.} \code{DivergenceAwarePatchingResult} extends
\code{PatchingResult} with
\code{divergence\_info}: list of
\code{DivergenceInfo}(component, divergence\_score, is\_divergent,
divergence\_type, confidence),
\code{has\_pernicious\_divergence},
\code{reliability\_score},
\code{divergent\_components}.

\subsection{Misalignment cascade detector}

\paragraph{Underlying theory.} Natural Emergent Misalignment from
Reward Hacking (MacDiarmid et al., 2025) reports that reward-hacked policies exhibit
correlated failures across superficially distinct misalignment
dimensions, suggesting that several visible failure modes are driven
by a shared underlying mechanism. Detection at reward-model level
helps triage which behaviours to test for in the policy.

\paragraph{Operationalisation.} The detector ships six dimensions
(alignment-faking, malicious-cooperation, sabotage,
self-preservation, deception, sycophancy), each with default test
pairs of the form (prompt, aligned response, misaligned response).
Per-dimension misalignment is the mean preference for the misaligned
response across the pairs of that dimension. The pairwise correlation
matrix across dimensions is computed; pairs with $|r| > $ threshold
(default 0.5) are reported as correlated; connected components in the
above-threshold correlation graph (union-find) are reported as
clusters. The primary failure mode is the dimension maximising
$\sum_{d' \neq d} |r(d, d')|$.

\paragraph{Output.} \code{CascadeReport} with fields
\code{dimensions\_tested},
\code{per\_dimension\_scores},
\code{correlation\_matrix},
\code{cascade\_risk\_score} =
$0.4 \min(1, \bar{m}/0.2) + 0.3 \bar{|r|} + 0.3 f_\mathrm{correlated}$,
\code{correlated\_pairs},
\code{cascade\_clusters},
\code{primary\_failure\_mode}, \code{recommendations}. A
\code{cross\_validate\_with\_hacking(hacking\_report, cascade\_report)}
helper computes correlations between cascade dimensions and
hacking-detector dimensions for joint analysis.

\paragraph{Caveat in implementation.} With shipped sample sizes (1--3
pairs per dimension) the correlation matrix is degenerate: any
non-trivial pair of length-two sequences correlates at $\pm 1$. We
return the score as directional triage and flag the small-sample
caveat in the report's \code{recommendations} field.

\subsection{Reward-term conflict analyser}

\paragraph{Underlying theory.} Aligned, Orthogonal or In-Conflict
(Kaufmann et al., 2026) argues that for two reward terms the geometry of their
direction vectors --- aligned, orthogonal, or in-conflict ---
determines whether joint optimisation is safe; in-conflict terms
incentivise the policy to hide its reasoning to maximise the
combined reward.

\paragraph{Operationalisation.} The analyser accepts term directions
either supplied directly or learned from contrastive activations.
\code{learn\_term\_directions(term\_pairs)} takes per-term contrastive
pairs (prompt, preferred-on-this-term, dispreferred-on-this-term),
collects activations at the final layer, computes
$\delta_p = h_\mathrm{pref} - h_\mathrm{dispref}$, averages across
pairs, and normalises:
$d_t = \overline\delta_t / \lVert \overline\delta_t \rVert$. Pairwise
classification thresholds: $\cos > 0.5$ aligned;
$|\cos| < 0.2$ orthogonal; $\cos < -0.3$ in-conflict; otherwise
classified by sign.

\paragraph{Output.} \code{ConflictReport} with
\code{term\_names},
\code{term\_directions},
\code{pairwise\_analysis} (per pair: cosine similarity,
relationship, conflict severity, recommendations),
\code{relationship\_matrix},
\code{similarity\_matrix},
\code{in\_conflict\_pairs},
\code{overall\_conflict\_score},
\code{monitorability\_risk}.
For multi-objective heads (ArmoRM)
\code{analyze\_multi\_objective\_model(objective\_names)} is the
direct entry point; it pulls the per-objective directions from the
adapter and runs the same analysis.

\subsection{Concept-vector analysis}

\paragraph{Underlying theory.} Chen et al.\ (2025) extract concepts as
linear directions in residual-stream activations and demonstrates
that adding such a direction at inference time produces predictable
behavioural changes. The concept-vector analyser ports this
methodology to reward models: a concept direction's reward alignment
is its cosine with $w_r$, and its causal effect is the slope of the
reward under residual-stream addition.

\paragraph{Extraction.} \code{extract\_concepts(concept\_pairs, layer)}
takes per-concept contrastive pairs (prompt, positive example,
negative example), computes
$\delta_p = h^{(\ell)}_\mathrm{pos} - h^{(\ell)}_\mathrm{neg}$ at the
chosen layer (default: final), averages, and normalises:
$v_c = \overline\delta_c / \lVert \overline\delta_c \rVert$. The norm
of the unnormalised average is reported as a separability score.

\paragraph{Alignment and dose-response.} For each concept,
$\mathrm{align}(c) = v_c^\top (w_r / \lVert w_r \rVert)$. The causal
dose-response curve is generated by adding $\alpha v_c$ into the
final-token residual at the chosen layer for
$\alpha \in \{-2, -1, -0.5, 0.5, 1, 2\}$ and recording the resulting
reward; the slope of $\Delta r$ on $\alpha$ is the causal slope
reported in Table~\ref{tab:concepts}. A linear concept that is also
linearly read out by $w_r$ produces a clean straight-line
dose-response, as Skywork's agreement concept does in \S5.7.

\paragraph{Output.} \code{ConceptAlignmentReport} with
\code{concepts}: list of \code{ConceptInfo}(name, direction,
reward\_alignment, mean\_activation\_positive,
mean\_activation\_negative, separability, is\_reward\_aligned,
hacking\_risk),
\code{reward\_aligned\_concepts}, \code{anti\_reward\_concepts},
\code{high\_risk\_concepts}, \code{overall\_hacking\_risk},
\code{recommendations}.
\code{quick\_concept\_analysis(model)} is a convenience that runs the
six default concepts and returns the report directly.

\paragraph{Worked interpretation.} On Skywork, the agreement concept
returns alignment 0.340 and slope $+1.002$: the reward signal
\emph{contains} a near-pure linear copy of the agreement direction.
Combined with the hacking detector's verdict that Skywork
\emph{penalises} sycophancy ($d = -5.82$), the picture is consistent:
the model rewards substantive agreement (correctness) but penalises
flattering agreement (sycophancy). The library makes this combined
reading available in two function calls.

\section{Reproducibility}
\label{app:repro}

All experiments use the released library against HuggingFace
\code{transformers} 4.46.x and PyTorch 2.5.x, in bfloat16 on a single
NVIDIA A100-80GB. Seeds: \code{torch.manual\_seed(0)},
\code{numpy.random.seed(0)}; \code{model.eval()} and
\code{torch.inference\_mode()} are used throughout; no model weights
are modified by any analysis.

All experiments are reproducible from the released library: load each
reward model via \code{RewardModel.from\_pretrained}, iterate over
RewardBench preference pairs, and call the relevant primitive
(\code{RewardLens.trace}, \code{ComponentAttribution.attribute},
\code{ActivationPatcher.patch\_all\_components},
\code{HackingDetector.scan}, \code{quick\_concept\_analysis}) on each
pair. Per-pair outputs are flat dataclasses; aggregation across a
dataset is plain Python.

Per-pair crystallisation depths, per-component attribution and
patching outputs, and per-pair hacking deltas are written to a flat
JSON summary; reproducing the paper's tables from that file is a
trivial aggregation.

\section{Compact API reference}
\label{app:api}

The user-facing API, organised by purpose. Every entry is reachable
from the package root (\code{from reward\_lens import ...}).

\begin{lstlisting}
# Model wrapping
RewardModel.from_pretrained(name, ...)
    .reward_direction, .reward_bias
    .d_model, .n_layers, .n_heads, .d_head
    .score(prompt, response)
    .score_pair(prompt, pref, dispref)
    .forward_with_cache(prompt, response, cache_full_sequences=False)
    .forward_with_cache_from_inputs(input_ids, attn_mask, ...)
    .project_onto_reward(hidden_state)
    .hooks() -> context manager
ModelAdapter (LlamaAdapter | MistralAdapter | Gemma2Adapter |
              ArmoRMAdapter | GenericAdapter)
get_adapter(model, model_name)

# Observational primitives
RewardLens(model).trace(prompt, pref, dispref) -> RewardLensResult
    .trace_single(prompt, response) -> RewardLensResult
reward_lens_plot(model, prompt, pref, dispref, save_path=None)
ComponentAttribution(model).attribute(prompt, pref, dispref) -> ComponentResult
    .attribute_single(prompt, response) -> ComponentResult
TopKSAE(d_model, n_features, k)
ActivationCollector, SAETrainer, FeatureAnalyzer

# Causal primitives
ActivationPatcher(model)
    .patch_all_components(prompt, pref, dispref, mode="noising"|"denoising"|"zero")
    .patch_single_component(prompt, pref, dispref, layer_idx, component_type, mode)
DivergenceAwarePatching(model, distribution_estimator)
    .fit_distribution(prompts, responses)
    .patch_with_divergence_check(prompt, pref, dispref, divergence_threshold=2.0)
    .constrained_patch(...)

# Vulnerability and theory-grounded extensions
HackingDetector(model).scan(tests=[...], prompt=..., response=...) -> HackingReport
    .test_custom_pair(prompt, neutral, biased) -> BiasTestResult
DistortionAnalyzer(model).compute_distortion_index(...) -> DistortionReport
    .analyze_agentic_amplification(...) -> DistortionReport
    .compare_evaluation_strategies(...) -> dict[str, DistortionReport]
MisalignmentCascadeDetector(model).detect_cascade(...) -> CascadeReport
    .cross_validate_with_hacking(hacking_report, cascade_report)
RewardConflictAnalyzer(model).analyze_conflicts(reward_terms, ...) -> ConflictReport
    .learn_term_directions(term_pairs)
    .analyze_multi_objective_model(objective_names) -> ConflictReport
ConceptExtractor(model).extract_concepts(concept_pairs, layer)
    .analyze_reward_alignment(concept_vectors, alignment_threshold, hacking_concepts)
    .intervene_on_concept(prompt, response, concept_vector, strength, layer)
quick_concept_analysis(model) -> ConceptAlignmentReport

# Comparative
Comparator([model_a, model_b, ...]).compare(prompt, pref, dispref) -> ComparisonResult

# Curated data
diagnostic_data.get_diagnostic_pairs(dimensions=[...])
diagnostic_data.get_all_prompts_and_responses()
\end{lstlisting}

\end{document}